\documentclass{bvm}

\begin{document}

\newif\ifBlindPediatric
\BlindPediatricfalse  

\newcommand{\BlindedPediatric}[1]{%
  \ifBlindPediatric\textbf{[blinded for review]}\else#1\fi}

\newcommand{\bvmyear}{2026}

\selectlanguage{english} 

\title{From Adult Renal Masses to Pediatric Wilms Tumors}

\subtitle{Evaluating Cross-Domain Generalization of Kidney Lesion Segmentation on MRI}

\titlerunning{From Adult Renal Masses to Pediatric Wilms Tumors}

\ifBlindPediatric
  \author{\fname{Anonymous} \lname{Authors}\inst{1}}
  \authorrunning{Anonymous}
  \institute{ \inst{1} Blinded for Review }
  \email{author@institution.com}
\else
\author{
	\fname{Sarah} \lname[0000-0001-5184-4340]{de Boer} \inst{1}\equalContribution \affiliation{Radboudumc} \authorsEmail{sarah.deboer@radboudumc.nl}  \isResponsibleAuthor,
    \fname{Myrthe A. D.} \lname[0000-0003-0640-6434]{Buser} \inst{1}\equalContribution,
    \fname{Hartmut} \lname[0000-0002-6204-571X]{Häntze} \inst{1,2}\equalContribution \affiliation{Radboudumc},
    \fname{Jawed} \lname[]{Nawabi} \inst{3},
    \fname{Lisa C.} \lname[]{Adams} \inst{4},
    \fname{Alida F. W.} \lname[]{van der Steeg} \inst{5}, 
    \fname{Marry M.} \lname[]{van den Heuvel-Eibrink} \inst{5,6},
    \fname{Annemieke S.} \lname[]{Littooij} \inst{5,7},
    \fname{Keno K.} \lname[0000-0001-9249-8624]{Bressem} \inst{4,8,9,10},
    \fname{Mathias} \lname[0000-0001-8157-8055]{Prokop} \inst{1},
    \fname{Alessa} \lname[0000-0002-7602-803X]{Hering} \inst{1}
}

\authorrunning{de Boer, Buser, Häntze et al.}

\institute{
\inst{1} Department of Medical Imaging, Radboudumc, Nijmegen, The Netherlands\\
\inst{2} Department of Radiology, Charité - Universitätsmedizin Berlin, Berlin, Germany \\
\inst{3} Department of Neuroradiology, Charité - Universitätsmedizin Berlin, Berlin, Germany \\
\inst{4} Department of Diagnostic and Interventional Radiology, Klinikum rechts der Isar, TUM University Hospital, Technical University of Munich, Munich, Germany \\
\inst{5} Princess Máxima Center for Pediatric Oncology, Utrecht, The Netherlands  \\
\inst{6} Wilhelmina Children's Hospital, Division of Childhealth, Utrecht, The Netherlands \\
\inst{7} Department of Radiology and Nuclear Medicine, University Medical Center Utrecht, Utrecht, The Netherlands \\
\inst{8} National Center for Tumor Diseases West, Essen, Germany \\
\inst{9} Institute of Artificial Intelligence in Medicine, University Hospital Essen, Essen, Germany \\
\inst{10} Department of Diagnostic and Interventional Radiology and Neuroradiology, University Hospital Essen, Essen, Germany \\
}

\email{sarah.deboer@radboudumc.nl}
\fi

\maketitle

\begin{abstract}
    Deep learning segmentation models for pediatric oncology are often trained on small, single-center cohorts, whereas much larger datasets exist for adult diseases, such as kidney cancer. Whether models trained on adult renal lesions generalize to pediatric Wilms tumors remains unclear. We trained a segmentation model on single-sequence T1- or T2-weighted adult kidney MRI and evaluated it on a held-out adult test set and a pediatric Wilms tumor cohort, where it was compared with a pediatric-specific nnU-Net. Segmentation performance was quantified using the Dice similarity coefficient (DSC) and 95th-percentile Hausdorff distance. On the adult test set, the model achieved a median DSC of 0.86 for renal lesions. On the pediatric cohort, the adult-trained model achieved a median DSC of 0.85 compared with 0.90 for the pediatric-specific model. 
    Despite differences in age, anatomy, and tumor type, the adult-trained model showed promising performance on pediatric Wilms tumors, with only a modest difference in median DSC compared with the pediatric-specific model. However, non-inferiority could not be shown, highlighting the value of pediatric training data. In future work, a fine-tuning strategy could be explored. 
    
\end{abstract}

\section{Introduction}
Deep learning-based segmentation is increasingly used to support tumor volume assessment, treatment planning, and response monitoring in pediatric oncology. However, pediatric imaging datasets are inherently small because of the low incidence of pediatric cancers and the practical limitations in assembling multi-center datasets with expert annotations. Although Wilms tumor is the most common pediatric renal malignancy~\cite{spreafico2006}, research remains constrained by the scarcity of large, annotated imaging datasets~\cite{buser2023}, in contrast to the substantially larger cohorts available for adult renal diseases. 

Whether models trained on much larger adult datasets can generalize effectively to the pediatric setting remains an open question. For CT-based abdominal organ segmentation, Chatterjee et al.\ showed that an adult-trained model (TotalSegmentator ~\cite{wasserthal_2023}) achieves significantly lower Dice similarity coefficients (DSCs) on pediatric than on adult CT (0.73 vs.\ 0.81), with the largest performance drops in smaller anatomical structures~\cite{chatterjee2024}. These findings motivate investigating whether a similar generalization gap exists for adult-to-pediatric lesion segmentation on MRI. To the best of our knowledge, this has not been previously investigated. Moreover, the lack of a publicly available adult MRI model for kidney cancer segmentation has so far limited direct assessment of adult-to-pediatric transfer. 

In this work, we train a segmentation model on adult kidney lesion MRI and evaluate it on the pediatric Wilms tumor cohort previously used by Buser et al.~\cite{buser2023}. Specifically, we (1) quantify the generalization gap between the adult model's performance on an adult hold-out test set and its performance on pediatric data, and (2) directly compare the adult model with a pediatric-specific nnU-Net, on a per-case basis. To enable further evaluation of adult-to-pediatric transfer alongside existing public pediatric models~\cite{BUSER2026163151}, we release our adult model at \BlindedPediatric{\url{https://github.com/DIAGNijmegen/oncology-kidney-abnormality-segmentation}}.

\section{Materials and Methods}

\begin{figure}
    \centering
    \includegraphics[width=0.95\linewidth]{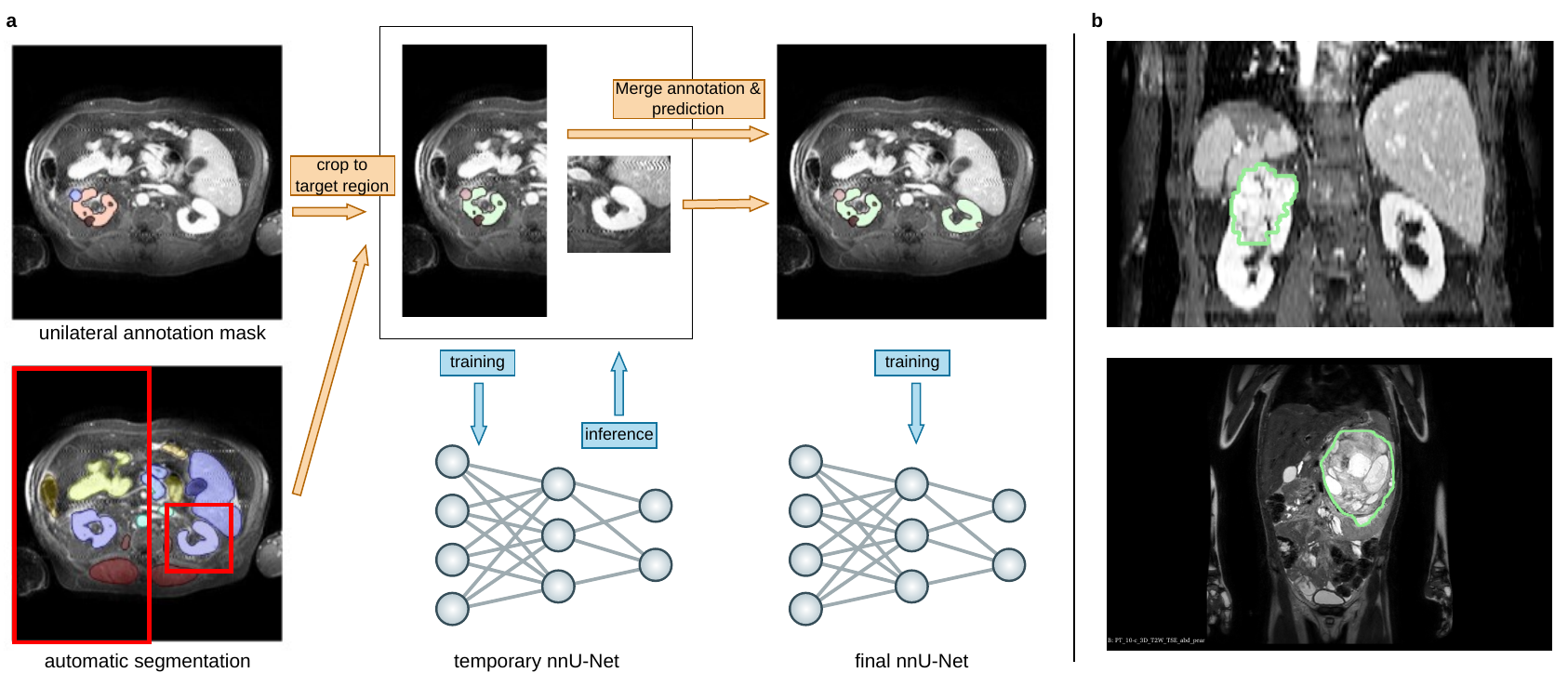}
    \caption{(a) Semi-supervised model training: MRSegmentator was used to localize both kidneys. A temporary nnU-Net was trained on manually annotated crops and used to generate pseudo-labels for previously unlabeled kidneys. Manual and automatic annotations were then merged to train the final model. (b) Tumor examples from the adult and pediatric cohorts.}
    \altText{The figure is split in two parts, "a" and "b". "A" shows the semi-supervised model training. MRSegmentator was used to localize both kidneys. A temporary nnU-Net was trained on manually annotated crops and used to generate pseudo-labels for previously unlabeled kidneys. Manual and automatic annotations were then merged to train the final model. "B" shows tumor examples from the adult and pediatric cohorts. Notably the adult scan is focuses on the abdominal regions and the kidney and tumors take a large space of the image. The pediatric scan, on the other side, appears like its zoomed out, as the body of the patient is much smaller. Hence, the complete upper torso from shoulders to pelvis, including both arms, is visible. Despite the comparable smaller size of the patients, both pediatric and adult tumor appear similarly large in absolute terms. }
    \label{fig:1}
\end{figure}

\subsection{Data}
\label{sec:data}

\paragraph{Adult cohort} The adult MRI data consisted of 1794 MRI sequences from 689 patients (age 38--74 years, 450 male) acquired at \BlindedPediatric{Charité – Universitätsmedizin Berlin}. It included cases of renal cell carcinoma, urothelial carcinoma, and oncocytoma, with T1-weighted, post-contrast fat-suppressed T1-weighted, and T2-weighted acquisitions. In previous work \BlindedPediatric{~\cite{haentze2024improve}}, a medical student annotated tumors, cysts and corresponding kidneys; kidneys without tumors remained unsegmented. In this work, we unified tumors and cysts to a more general class "renal mass" to reflect a single abnormality. Twenty percent of the data (138 patients, 359 sequences) were held out for in-domain (adult-on-adult) evaluation, with the remaining 80\% used for training. The institutional ethics board approved retrospective use of the dataset for research. Dataset characteristics are summarized in Table~\ref{XXXX-tab-datasets}.

\paragraph{Pediatric cohort} The pediatric test cohort consisted of MRI data acquired at \BlindedPediatric{Princess Máxima Center for Pediatric Oncology}, comprising 45 patients with Wilms tumor (age 7--109 months, 22 male, 52 tumors, 7 bilateral)~\cite{buser2023}. All scans were 3D T2-weighted acquisitions (1.5\ts T, voxel size = 0.83\ts mm, slice thickness = 1.15\ts mm), including manual reference segmentations of the lesions, annotated by technical physicians and checked by a radiologist.  
The institutional ethics board approved retrospective use of this dataset for research. Example images comparing adult to pediatric cases are shown in Figure \ref{fig:1}b.

\begin{table}[t]
\caption{Overview of the adult and pediatric cohorts. Renal mass volume is reported as the median and interquartile range.}
\label{XXXX-tab-datasets}
\begin{tabular*}{\textwidth}{p{3.0cm}@{\extracolsep\fill}p{3.8cm}p{3.8cm}}
\hline
 & Adult cohort & Pediatric cohort \\
\hline
Institution        & \BlindedPediatric{Charité -- Universitätsmedizin Berlin} & \BlindedPediatric{Princess Máxima Center for Pediatric Oncology} \\
Use in this study  & Training and in-domain test & External test (zero-shot) \\
Patients (sequences) & 689 (1794) & 45 (45) \\
\quad Training     & 551 (1435) & -- \\
\quad Test         & 138 (359)  & 45 (45) \\
Age range          & 38--74\ts years & 7--109\ts months \\
Sex (male)         & 450 (65\,\%) & 22 (49\,\%) \\
Diagnosis          & Renal cell carcinoma, urothelial carcinoma, oncocytoma, renal cysts & Wilms tumor \\
Sequences          & T1w, fat-suppressed post-contrast T1w, T2w & T2w \\
Annotated structures & Tumor-bearing kidney, solid masses, cysts & Tumors \\
Solid masses       & & \\
\quad Components   & 2208 & 52 \\
\quad Volume (ml) & 15.3 (3.9--65.8) & 200.2 (95.1--497.0) \\
Cysts              & & \\
\quad Components   & 2432 & -- \\
\quad Volume (ml) & 0.6 (0.2--2.4) & -- \\
\hline
\end{tabular*}
\end{table}

\subsection{Model training}
The reference annotations in the adult cohort only included kidneys and cysts ipsilateral to a solid tumor. This was insufficient for training a model, as the missing annotations for contralateral structures would affect the training loss. Consequently, we designed a semi-supervised training workflow to mitigate the issue of missing annotations, as depicted in Figure~\ref{fig:1}a.

First, we used the publicly available MRSegmentator model~\cite{my_hantze2025segmenting} to delineate contralateral kidneys. Notably, these delineations did not include potential renal masses and were therefore only used to calculate bounding boxes. For each MRI sequence, we computed two disjoint crops. The first enclosed the manual annotations while excluding the contralateral kidney. The second enclosed the contralateral kidney with a 3\ts cm margin. In cases where both kidneys were tumor-bearing, the reference annotations were used without cropping. Next, we trained a 3D nnU-Net-based segmentation network~\cite{isensee2021nnu} exclusively on the manually annotated kidney crops and used the resulting model to predict masks for the contralateral structures.

Lastly, we merged the reference annotations and predicted contralateral segmentations into unified masks and trained a new, independent nnU-Net on the complete sequences from the adult training partition. Notably, for the evaluation, we restricted our analysis to the crops around the manual annotations in the hold-out data. No fine-tuning or adaptation of any kind was performed on the pediatric data; all pediatric evaluations were strictly zero-shot.

\subsection{Experimental design and evaluation}
We addressed three research questions. First, we established an in-domain baseline by evaluating the adult model on the held-out adult test set. Second, we assessed cross-domain generalization by evaluating the same model on the pediatric Wilms tumor dataset. Third, we compared its performance in the pediatric setting with that of the pediatric-specific nnU-Net model developed by Buser et al.~\cite{buser2023}.

Segmentation performance was quantified using the Dice similarity coefficient (DSC) and 95th-percentile Hausdorff distance (HD95). Metrics were calculated for both kidney and renal mass classes in the adult dataset and for renal mass alone in the pediatric dataset.
On the pediatric dataset, non-inferiority of the adult model relative to the pediatric-specific model was assessed using a paired Wilcoxon signed-rank test on DSC, with a pre-specified non-inferiority margin of $\delta=0.03$ based on previously reported inter-reader variability in manual cyst annotations~\cite{sore_2024}. For adult-to-pediatric generalization, we did not perform significance testing because differences in tumor volumes, image sizes, and annotation protocols between datasets may independently affect segmentation metrics. Instead, we report these differences descriptively.

\section{Results}

\subsection{Adult in-domain performance}
We evaluated the adult MRI segmentation model on the adult hold-out test set, which consisted of 359 sequences from 138 patients. Across all sequence types, the model achieved a median kidney DSC of 0.92 (IQR: 0.87 -- 0.94), and a median HD95 of 3.5\ts mm (IQR: 2.8 -- 5.8). For renal mass segmentation, the median DSC was 0.86 (IQR: 0.79 -- 0.91) and the median HD95 was 6.3\ts mm (IQR: 4.0 -- 15.6). Performance varied by sequence type, with a median renal mass DSC of 0.87 for T1-weighted fat-suppressed sequences, 0.87 for T2-weighted sequences, and 0.84 for conventional T1-weighted sequences. 

\subsection{Generalization gap}
The adult model was evaluated on a pediatric cohort of 45 patients with Wilms tumor. It achieved a median tumor DSC of 0.85 (IQR: 0.64 - 0.89) and a median HD95 of 11.0\ts mm (IQR: 6.8 - 16.8).  This included 5 patients with a DSC of 0, of which in three cases the model did not produce any kidney or tumor segmentation (all in patients younger than 10 months). 

\subsection{Comparison with pediatric model}
\begin{figure}
    \centering
    \includegraphics[width=0.9\linewidth]{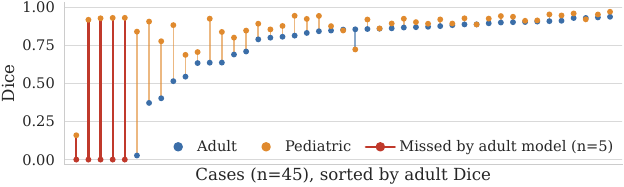}
    \caption{Case-level DSC comparison between the adult- and pediatric-trained models on the Wilms tumor cohort.}
    \label{fig:2}
    \altText{The figure shows the paired Dice Similarity Coefficients of both adult and pediatric model for every case of the pediatric data. For 5 cases, the adult model did not detect the tumors resulting in Dice values of 0, while the pediatric model achieved values of around 0.8. However, for many of the other cases, both models achieved similar results.}
\end{figure}

\begin{figure}
    \centering
    \includegraphics[width=\linewidth]{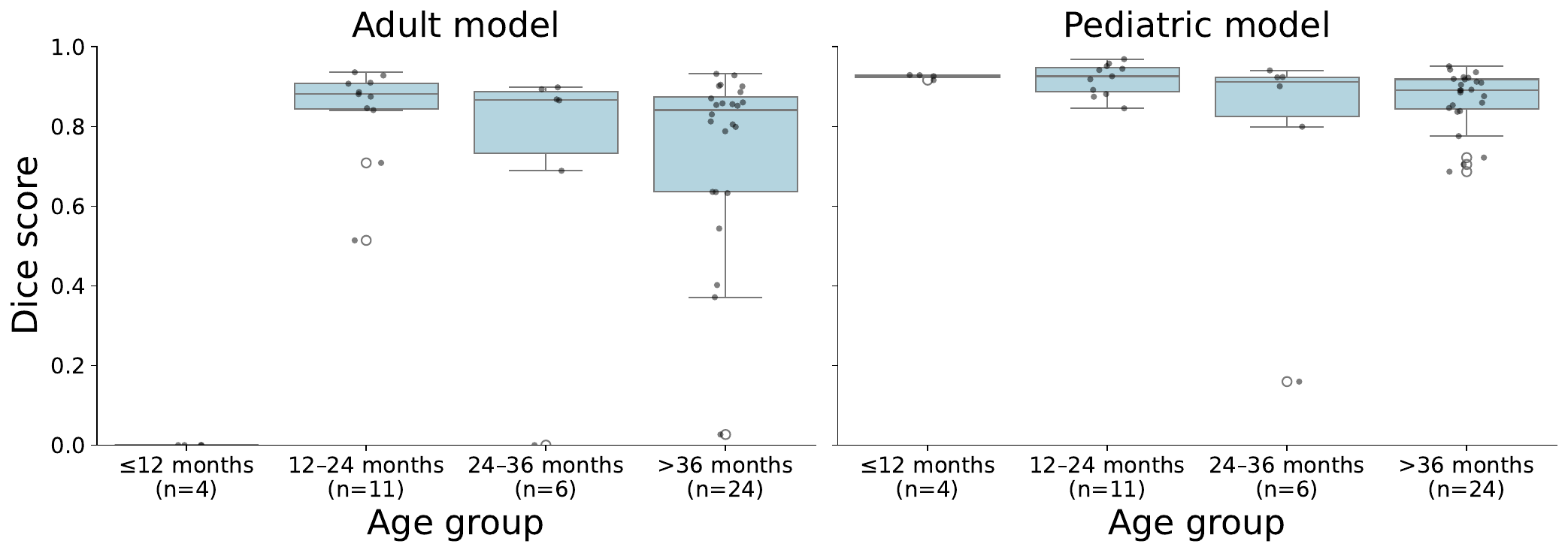}
    \caption{Subgroup analysis comparing the performance (in DSC) per patient age group for the adult (left) and pediatric (right) models.}
    \label{fig:3}
\end{figure}

On the pediatric dataset, the pediatric-specific model achieved a median tumor DSC of 0.90 (IQR: 0.86 -- 0.92), compared with 0.85 (IQR: 0.64 -- 0.89) for the adult model. The corresponding median HD95 values were 6.6\ts mm (IQR: 4.9 -- 11.8) and 11.0\ts mm (IQR: 6.8 -- 16.8), respectively. Non-inferiority of the adult model to the pediatric-specific model could not be demonstrated. A per-case DSC comparison is presented in Figure~\ref{fig:2}. In Figure~\ref{fig:3}, we present an analysis comparing the performance across patients' age groups for the adult (left) and pediatric (right) model.

\section{Discussion and conclusion}

In the adult cohort, the model achieved a median kidney DSC of 0.92. Although this is lower than the DSC of up to 0.95 reported for both kidneys by MRSegmentator~\cite{my_hantze2025segmenting}, direct comparison is limited because our evaluation was restricted to tumor-affected kidneys. For renal lesion segmentation, the model achieved a DSC of 0.86, which is comparable to previously reported performance on CT~\cite{deBoer2026RobustRenal}, and within the range of reported results for MRI~\cite{anari_2023,anusha_2025,pavarut_2025}.

Despite not being trained on Wilms tumors or pediatric data, the model generalized to the pediatric cohort, potentially benefiting from the broad "renal mass" class, which included multiple types of renal anomalies. However, performance was lower than that of the pediatric-specific model, with no tumor segmentation in four cases. This gap may reflect differences in both disease and anatomical scale. Wilms tumors differ morphologically from the renal lesions represented in the adult training data, and pediatric images contain substantially smaller anatomical structures. These scale differences may be particularly relevant given the model’s fixed isotropic resampling, which can reduce the effective resolution of small structures. We hypothesized that the performance of the adult-trained model would correlate with patient age, as MRI of older children should be more similar to the training data distribution. However, interestingly, the DSC performance of the adult model did not systematically increase with patient age.

This study has limitations. 
Most importantly, the comparison between the adult and pediatric models was not fully symmetric: the pediatric model was evaluated using cross-validation on the pediatric dataset, whereas the same dataset was used as an external test set for the adult model. This limits direct interpretation of the performance difference. Furthermore, the pediatric cohort was relatively small, and the datasets differed in acquisition characteristics and annotation protocols, making it difficult to attribute the observed performance difference to any single source of domain shift. 

Future work should investigate whether domain adaptation or targeted fine-tuning can improve adult-to-pediatric transfer while reducing the need for large pediatric training cohorts. Such approaches could be valuable for rare pediatric diseases, where assembling large annotated datasets is challenging.

Overall, the adult model demonstrated the ability to generalize to pediatric Wilms tumor MRI. However, the observed performance gap relative to a pediatric-specific model indicates that large adult datasets cannot fully compensate for pediatric-specific differences in anatomy, imaging, and disease presentation. Dedicated pediatric models therefore remain valuable, while the demonstrated transferability of the adult model suggests a potential role for large adult datasets in supporting pediatric annotation and model development.

\begin{acknowledgement}
    \BlindedPediatric{This research is funded by the European Union under HORIZON-HLTH-2022: COMFORT (101079894). Views and opinions expressed are however those of the author(s) only and do not necessarily reflect those of the European Union or European Health and Digital Executive Agency (HADEA). Neither the European Union nor the granting authority can be held responsible for them.}
\end{acknowledgement}

\printbibliography

\end{document}